\documentclass[letterpaper, 10 pt, conference]{ieeeconf}

\IEEEoverridecommandlockouts
\usepackage{comment}
\usepackage{cite}
\usepackage{amsmath}
\usepackage{amsfonts}
\usepackage{mathtools}
\usepackage{amssymb}
\usepackage{tabularx}
\usepackage{multirow}
\usepackage{graphicx}
\usepackage{array}
\usepackage{bm}
\usepackage{hhline}
\title{\LARGE \bf
Joint On-Manifold Gravity and Accelerometer Intrinsics \\ Estimation for Inertially Aligned Mapping
}

\author{Ryan Nemiroff, Kenny Chen, and Brett T. Lopez%
\thanks{The authors are with the Verifiable and Control-Theoretic Robotics Laboratory, University of California Los Angeles, Los Angeles,
CA, USA {\tt\footnotesize \{ryguyn, kennyjchen, btlopez\}@ucla.edu}.}}

\newcommand{\defeq}{\vcentcolon=}
\newcommand{\I}{\mathbf{I}}
\newcommand{\R}{\mathbf{R}}

\newcommand{\id}[0]{{\bm{\varepsilon}}}
\newcommand{\manEl}[0]{\mathbf{d}}
\newcommand{\tanEl}[0]{\mathfrak{d}}
\newcommand{\tanB}[0]{\mathrm{B}}

\begin{document}
\bstctlcite{IEEEexample:BSTcontrol}
\maketitle
\thispagestyle{empty}
\pagestyle{empty}


\begin{abstract}
Aligning a robot's trajectory or map to the inertial frame is a critical capability that is often difficult to do accurately even though inertial measurement units (IMUs) can observe absolute roll and pitch with respect to gravity.
Accelerometer biases and scale factor errors from the IMU's initial calibration are often the major source of inaccuracies when aligning the robot's odometry frame with the inertial frame, especially for low-grade IMUs. 
Practically, one would simultaneously estimate the true gravity vector, accelerometer biases, and scale factor to improve measurement quality but these quantities are not observable unless the IMU is sufficiently excited.
While several methods estimate accelerometer bias and gravity, they do not explicitly address the observability issue nor do they estimate scale factor. 
We present a fixed-lag factor-graph-based estimator to address both of these issues. 
In addition to estimating accelerometer scale factor, our method mitigates limited observability by optimizing over a time window an order of magnitude larger than existing methods with significantly lower computational burden. The proposed method, which estimates accelerometer intrinsics and gravity separately from the other states, is enabled by a novel, velocity-agnostic measurement model for intrinsics and gravity, as well as a new method for gravity vector optimization on $S^2$. 
Accurate IMU state prediction, gravity-alignment, and roll/pitch drift correction are experimentally demonstrated on public and self-collected datasets in diverse environments.
\end{abstract}


\section{Introduction}
Inertial measurement units (IMUs) have become a core sensor in state-of-the-art odometry estimation and SLAM systems because they provide high-rate measurements that are complimentary to other sensing modalities, such as vision or LiDAR.
For instance, incorporating an IMU into a LiDAR odometry algorithm can help correct for motion-induced distortion of the point cloud while also reducing the computational complexity of scan-matching by providing a reasonable initial guess to the nonlinear solver \cite{zhang2014loam, shan2020lio, xu2022fast, dlio}.
One can also align geometric or semantic maps with the inertial frame by taking advantage of the IMU's ability to observe gravity. 
However, raw IMU measurements are corrupted by accelerometer and gyroscope biases in addition to inaccurate scale factors that severely degrade measurement quality. 
While gyroscope bias estimation is fairly straightforward, accurate accelerometer bias estimation can only be achieved when the gravity vector is known relative to each accelerometer measurement. 
One way this can be achieved is by simultaneously estimating the gravity vector along with accelerometer biases so that both estimates eventually converge to their true values. 
Yet, several state-of-the-art LIO methods \cite{shan2020lio, dlio} are lacking this capability and assume during initialization that either the robot is aligned with the inertial frame or that the biases are negligible.
This assumption not only introduces errors in the biases used by the algorithm --- which subsequently affects pose estimates --- but it also prevents proper alignment of the resulting trajectory and map to the inertial frame.

\begin{figure}
    \centering
    \includegraphics[width=0.99\columnwidth]{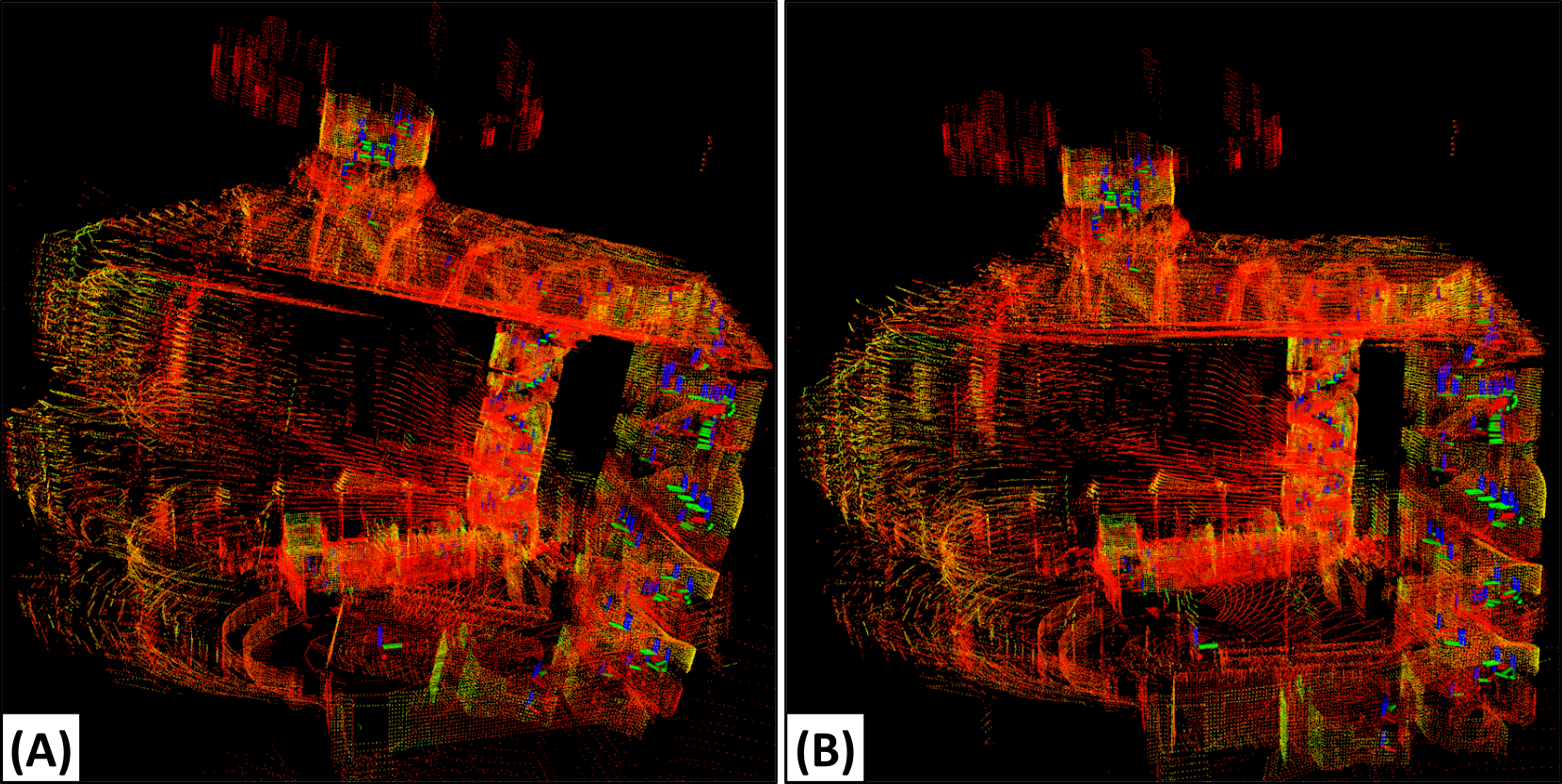}
    \vspace{-6mm}
    \caption{Maps produced by DLIO \cite{dlio} without (A) and with (B) global map gravity correction on the Hilti-Oxford Cupola dataset \cite{zhang2023hiltioxford}. Our method is able to compensate for extreme drift in odometry that may occur in challenging environments by estimating gravity and adding costs to align the map. As a result, the map in (B) is more accurate and self-consistent in roll and pitch.}
    \label{fig:map}
    \vskip -0.2in
\end{figure}


For the reasons outlined above, several recent LIO methods perform concurrent gravity estimation \cite{qin2020lins, xu2022fast, wang2022dliom}, with a few incorporating this estimate as a penalty in the back-end global map optimizer \cite{ye2019tightly, ramezani2022wildcat}.
Despite these efforts, there still remain challenges in estimating accelerometer bias and gravity in real-time. 
First, bias and gravity are only fully observable under certain conditions. 
That is, a nonzero angular velocity which changes axis of rotation at some point is required. A stronger condition, uniform complete observability, requires that the axis of rotation changes sufficiently frequently~\cite{batista2011obs}. 
However, a typical robot may not meet this criteria for long stretches of time. 
A robot which undergoes frequent yaw rotation will achieve a greater degree of observability compared to no rotation at all, but will not achieve full observability. 
This means that the bias and gravity estimates may not converge for a significant time after robot initialization, and if drift occurs the new values may not be immediately observable. 
Since measurements which achieve the necessary excitation may occur far apart from each other, an estimator can benefit from explicitly optimizing over a long history of measurements. 
Second, scale factor and cross-axis sensitivity errors can become significant for consumer-grade IMUs. 
These intrinsic IMU properties are rarely considered in existing LIO methods.



In this work, we address the aforementioned challenges by deriving a velocity-agnostic measurement model for accelerometer intrinsics and gravity that constrains these variables in a factor graph, separate from the other state variables. 
This factor is optimized over measurements spanning a sufficiently large time window to concurrently estimate accelerometer bias, scale factor, cross-axis sensitivity, and gravity direction while maintaining low computational overhead. 
Since the new estimator only includes slow-changing quantities, it becomes computationally tractable to optimize over a large time window. 
The approach is general and can be combined with any odometry or SLAM method, although in this work we will demonstrate the approach within a LiDAR odometry pipeline.
We introduce the following innovations to make this estimator possible:
\begin{enumerate}
    \item A novel velocity-agnostic measurement model for accelerometer intrinsics (sensitivity and bias) and gravity using pose and IMU measurements.
    \item A new method for gravity vector optimization over the 3D unit sphere $S^2$ within iSAM2~\cite{kaess2011isam2}.
    \item A factor graph incorporating the above to estimate accelerometer intrinsics and gravity.
\end{enumerate}
We demonstrate that, compared to an estimator which excludes gravity, better IMU prediction is achieved. Furthermore, gravity vector estimates are competitive with existing state-of-the-art methods. Finally, lower roll/pitch drift is seen by incorporating the gravity estimates into global map optimization (Fig.~\ref{fig:map}).


\section{Related Work}



LiDAR odometry (LO) systems primarily operate by aligning successive point clouds by means of solving a nonlinear least-squares problem. This alignment, otherwise known as \textit{point cloud registration}, outputs a best-fit homogeneous transformation which minimizes the error between corresponding points and/or planes between the two scans after converging to a local minimum. Incorporating an IMU can aid in system performance by correcting for motion distortion~\cite{zhang2014loam, shan2020lio, xu2022fast, dlio}, providing an initial guess for scan-matching~\cite{zhang2014loam, ramezani2022wildcat, shan2020lio, chen2022direct, dlio}, and recently, estimating and using gravity relative to a robot's body frame for improved bias estimation and localization~\cite{xu2022fast, ramezani2022wildcat, wang2022dliom, qin2020lins, kubelka2022gravity}.

Estimating gravity and accelerometer bias \textit{jointly} is crucial for either to be accurate,
and for constructing globally-aligned maps. For example, if biases are assumed to be less than $0.5\text{ m}/\text{s}^2$ and their estimates are initialized to $0$, then static gravity calibration can be off by up to $4^\circ$. While many works naively assume a constant gravity vector~\cite{dlio, shan2020lio, dai2020lio, xie2021lio}, there has been recent interest in constructing a more accurate representation by augmenting a system's state estimator to include gravity as a free variable. One popular approach for estimating the gravity vector is to use a probabilistic filter that combines accelerometer measurements and \textit{a priori} knowledge of the local gravity vector. For example, LINS~\cite{qin2020lins} employs a unique iterated error-state Kalman filter with body gravity as part of the local state, while FAST-LIO2~\cite{xu2022fast} similarly use an iterated extended Kalman filter to estimate gravity in the world frame. Wildcat~\cite{ramezani2022wildcat} and D-LIOM~\cite{wang2022dliom} instead estimate gravity as part of a local trajectory optimization technique, although they fail to estimate biases and gravity in a truly coupled manner.

Even methods that estimate gravity may not implement global map correction. This has only been seen in a few works, including Wildcat~\cite{ramezani2022wildcat}, which adds a gravity factor to each node in the global pose graph similar to our implementation. NeBula~\cite{agha2022nebula} does this as well, but only for poses where the robot is stationary, since an online gravity estimate is not available. Additionally, \cite{ye2019tightly} is unique in that it employs a popular fixed-lag trajectory optimization method based on IMU pre-integration~\cite{forster2016manifold}, but explicitly allows freedom of the local map's orientation; the gravity-informed orientations are then added as constraints to the global pose graph.

While some works intelligently handle gravity, none mentioned take into account accelerometer intrinsics beyond just bias (i.e., scale factor and cross-axis sensitivity). Most also do not estimate errors in the provided LiDAR-IMU extrinsics. Existing efforts for estimating these intrinsics/extrinsics include
either a more comprehensive initialization process during system startup~\cite{zhu2022robust} or
an offline calibration procedure~\cite{lv2022calib}. However, a calibration procedure or initialization with sufficient excitation may not always be available.

To this end, we present a system which accurately estimates accelerometer intrinsics (e.g., bias, scale factor, and cross-axis sensitivity) and gravity online and can be incorporated into any existing odometry framework in order to improve localization and mapping performance. Our approach is motivated by the need for accurate state estimation even when system excitation is limited and when accelerometer scale factor, cross-axis sensitivity, and rotational misalignment are non-negligible (where \textit{extrinsic} misalignment can additionally be thought of as part of the intrinsics). Unlike previous methods, we jointly optimize accelerometer instrinsics and gravity over a large time window and include more comprehensive intrinsics estimation in order to reduce error in the bias and gravity estimates.

\section{Methods}

\subsection{Overview}

The key developments for estimating accelerometer intrinsics and gravity are as follows. First, we derive a measurement from pose and acceleration observations which provides information about the sensitivity, bias, and gravity. No single measurement provides complete information for these variables, but a combination of measurements under sufficient excitation may achieve this. These measurements are added to a factor graph and are referred to as ``odometry factors."
Our factor graph encompasses a fixed-lag window which is divided evenly into a number of time intervals, each with its own set of variables. The oldest set of variables is marginalized out as it passes out of the lag window. Here we see the advantage of estimating values which only change on large time scales---far fewer variables are present in the graph in a given time window compared to the case where robot states are being estimated. For example, we may assume our variables to be constant over each 3 second window, and thus we only execute marginalization once every 3 seconds. The lightweight factor graph means we can optimize over a large fixed-lag window (e.g., one minute) with low computational overhead. One can also choose how often to recompute the estimates to manage the computation time.
We also add a ``gravity factor" for each keyframe in the global map, which describes where the gravity vector should point in the body frame.
Crucially, the map is allowed to shift to better align inertially because this new factor includes the covariance of the gravity estimate computed by the previous factor graph.
While we focus on keyframe-based odometry, this method could be adapted to other odometry systems as well.



\subsection{Velocity-Agnostic Odometry Factor} \label{sec:o_factor}
In this section, we derive a model to represent measurements of accelerometer intrinsics and gravity without the need of accurate translational velocity measurements.
Consider the accelerometer model
\begin{align}
\tilde{\mathbf{a}}_\mathcal{B}(t) = \mathbf{S}^{-1}\left(\R^{-1}(t)\left(\mathbf{a}_\mathcal{I}(t)+\mathbf{g}_\mathcal{I}\right) + \mathbf{b} + \mathbf{w}_a(t)\right) ,
\label{eq:accel}
\end{align}
where $\tilde{\mathbf{a}}_\mathcal{B}(t) \in \mathbb{R}^3$ is the known accelerometer measurement in the body frame ($\mathcal{B}$), $\mathbf{a}_\mathcal{I}(t)$ is the true acceleration in the inertial frame ($\mathcal{I}$), $\R(t)$ is the orientation of the robot represented as a rotation matrix, and $\mathbf{w}_a(t)$ is random noise assumed to have noise characteristics $\mathcal{N}(0,\,\sigma_a^2\I)$. The values $\mathbf{g}_\mathcal{I}$, $\mathbf{b}$, and $\mathbf{S} \approx \I_{3 \times 3}$ are gravity, accelerometer bias, and the accelerometer ``sensitivity matrix," respectively, which are assumed to be constant in the short term.
We represent the effects of scale factor for each axis, cross-axis sensitivity, and axis misalignment \cite{skog2006calibration, ang2004accel} as a single matrix $\mathbf{S}$. The sensitivity matrix also capture errors in the IMU rotation extrinsic which are indistinguishable from axis misalignment.

The convenience of the particular model in (\ref{eq:accel}) becomes clear when isolating $\mathbf{a}_\mathcal{I}(t)$, seen in (\ref{eq:accel_inv}). In particular, if noise is negligible and $\R(t)$ and $\tilde{\mathbf{a}}_\mathcal{B}(t)$ are assumed known, then $\mathbf{a}_\mathcal{I}(t)$ is linear in $\mathbf{S}$, $\mathbf{b}$, and $\mathbf{g}_\mathcal{I}$, such that
\begin{align}
\mathbf{a}_\mathcal{I}(t) = \R(t)\left(\mathbf{S}\tilde{\mathbf{a}}_\mathcal{B}(t) - \mathbf{b}\right) - \mathbf{g}_\mathcal{I} - \R(t)\mathbf{w}_a(t) \,.
\label{eq:accel_inv}
\end{align}
An odometry algorithm provides position estimates which are related to accelerometer measurements via integration. Let $\mathbf{p}$ and $\mathbf{v}$ denote position and velocity in the inertial frame. Given a set of sampled accelerations $\mathbf{a}_\mathcal{I}^0,\dots,\mathbf{a}_\mathcal{I}^n$ at times $t^0,\dots,t^n$ and positions $\mathbf{p}^0$, $\mathbf{p}^k$, $\mathbf{p}^n$ where $0 < k < n$,
\begin{align}
\begin{split}    
\mathbf{p}^k-\mathbf{p}^0 &= \beta_A\mathbf{v}^0 + \sum_{i=0}^k{\alpha_A^i\mathbf{a}_\mathcal{I}^i} \\
\mathbf{p}^n-\mathbf{p}^0 &= \beta_B\mathbf{v}^0 + \sum_{i=0}^n{\alpha_B^i\mathbf{a}_\mathcal{I}^i}
\end{split} \label{eq:two_intervals}
\end{align}
where the $\alpha$ and $\beta$ denote the proper coefficients resulting from integration, determined by the times $t^0,\dots,t^n$. That is, $\alpha$ has units $[\text{s}^2]$ and $\beta$ has units $[\text{s}]$. The reason for considering these two intervals is to eliminate $\mathbf{v}^0$, e.g.,
\begin{multline}
\frac{1}{\beta_B}(\mathbf{p}^n-\mathbf{p}^0) - \frac{1}{\beta_A}(\mathbf{p}^k-\mathbf{p}^0) \\
 = \frac{1}{\beta_B}\sum_{i=0}^n{\alpha_B^i\mathbf{a}_\mathcal{I}^i} - \frac{1}{\beta_A}\sum_{i=0}^k{\alpha_A^i\mathbf{a}_\mathcal{I}^i},
\label{eq:v_agnostic}
\end{multline}
which becomes useful once we substitute in expressions for the positions and accelerations in terms of their measurements. We assume a simple model for position measurements generated by, e.g., LiDAR point cloud registration:
\begin{align}
    & \tilde{\mathbf{p}}(t) = \mathbf{p}(t) + \mathbf{w}_p(t), & \mathbf{w}_p(t) \sim \mathcal{N}(0,\,\Sigma_p(t)) \,,
    \label{eq:point_measurement}
\end{align}
\noindent where the noise $\mathbf{w}_p(t)$ is independent for each measurement.
Importantly, we obtain a covariance matrix $\Sigma_p(t)$ for each position measurement which should be a sufficient representation of the actual uncertainty. In this work, we estimate the covariance based on the Hessian of GICP-based point cloud registration, but this is only a rough approximation. Computing covariance from point cloud registration is a problem of ongoing discussion \cite{brossard2020new}, but it is important here since poor geometry and weather conditions can significantly affect the observability of point cloud registration. The estimator should be aware of this to ensure the best possible accuracy. While the estimator may not be updated when LiDAR or vision quality is very poor, it provides a benefit in the previously estimated intrinsics, which are depended upon to maintain odometry estimates. 

Substituting (\ref{eq:accel_inv}) and {(\ref{eq:point_measurement}) for $\mathbf{a}_\mathcal{I}$ and $\mathbf{p}$ into (\ref{eq:v_agnostic}), we get
\begin{align}
    \mathrm{M}_{\mathrm{s}}\mathbf{s} + \mathrm{M}_{\mathrm{b}}\mathbf{b} + \mathrm{M}_{\mathrm{g}}\mathbf{g}_\mathcal{I} - \left(\gamma_1\tilde{\mathbf{p}}^0 + \gamma_2\tilde{\mathbf{p}}^k + \gamma_3\tilde{\mathbf{p}}^n\right) = \mathbf{w} \,,
    \label{eq:o_factor}
\end{align}
where the sensitivity matrix $\mathbf{S}$ is now represented as a vector $\mathbf{s} \in \mathbb{R}^9$. $\mathrm{M}_\mathrm{s}$, $\mathrm{M}_{\mathrm{b}}$, and $\mathrm{M}_{\mathrm{g}}$ are constant matrices dependent on the known $\R(t)$, $\tilde{\mathbf{a}}_\mathcal{B}(t)$, and $t^0,\dots,t^n$. 
The scalars $\gamma_1$, $\gamma_2$, and $\gamma_3$ are shorthand for ${\beta^{-1}_A} - \beta^{-1}_B$, $-{\beta^{-1}_A}$, and ${\beta^{-1}_B}$, respectively. 
The right-hand side of (\ref{eq:o_factor}) is a noise $\mathbf{w}$ determined by the measurement noises according to
\begin{align}
    \mathbf{w} = \sum_{i=0}^n{\alpha_C^i\R^i\mathbf{w}_a^i} - \left(\gamma_1\mathbf{w}_p^0 + \gamma_2\mathbf{w}_p^k + \gamma_3\mathbf{w}_p^n\right)
    \label{eq:o_factor_noise}
\end{align}
where the $\alpha_C^i$ are are coefficients derived from the $\alpha_A^i$, $\alpha_B^i$, $\beta_A$, and $\beta_B$. As a sum of independent Gaussian random variables, $\mathbf{w}$ is characterized by $\mathcal{N}(0,\Sigma)$, where the covariance is computed via
\begin{align}
    \Sigma = \left(\sum_{i=0}^n{\left(\alpha_C^i\right)^2}\right)\I + (\gamma_1)^2\Sigma_p^0 + (\gamma_2)^2\Sigma_p^k + (\gamma_3)^2\Sigma_p^n \,.
    \label{eq:o_factor_cov}
\end{align}
Noting that $\mathbb{E}\left[\mathbf{w}\right] = 0$, (\ref{eq:o_factor}) and (\ref{eq:o_factor_cov}) are sufficient to describe a measurement on $\mathbf{S}$, $\mathbf{b}$, and $\mathbf{g}_\mathcal{I}$ for a factor graph. In our factor graph, a new factor is added each time a pose measurement is received. This factor is based on the three most recent pose measurements, which correlate to $\tilde{\mathbf{p}}^0$, $\tilde{\mathbf{p}}^k$, and $\tilde{\mathbf{p}}^n$ in (\ref{eq:o_factor}).

A few remarks are in order. First, for the sake of clarity, equation (\ref{eq:two_intervals}) and onward are presented assuming simultaneity between pose and accelerometer measurements, but the actual implementation does not require this.
Additionally, the assumption that the robot orientation $\R(t)$ is known at all times is not a valid one; there will be errors in this measurement due to errors in point cloud registration and gyroscope integration. However, the decoupling of rotation and translation makes the estimation problem much simpler.
Therefore, one should take care to produce a sufficient estimate of the robot's orientation (e.g., via a geometric observer~\cite{dlio}). In our case, we typically see the state estimate before an observer update deviate from the incoming LiDAR pose by less than $1^\circ$, but further improvement may be achievable.
Finally, in our method, we take $\tilde{\mathbf{p}}^0$, $\tilde{\mathbf{p}}^k$, and $\mathbf{p}^n$ to be consecutive position measurements, but it is not clear that this is the best choice. In fact, there may be superior methods for constraining accelerometer instrinsics and gravity (i.e., leaving the initial velocity as a free variable).

\subsection{Gravity Optimization on $S^2$} \label{sec:s2}

\begin{figure}
    \centering
    \vspace{2mm}
    \includegraphics[width=0.9\columnwidth]{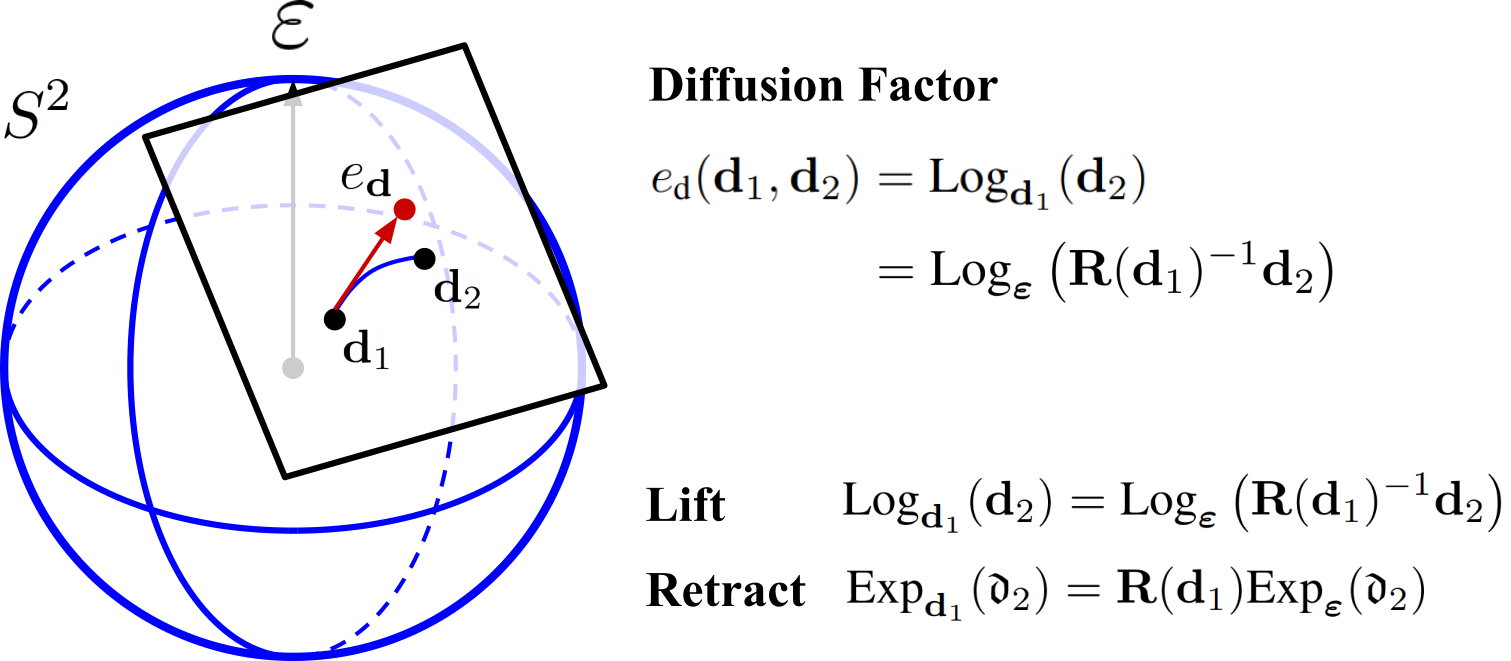}
    \vspace{-1mm}
    \caption{The $S^2$ manifold is a unit sphere which we use to construct functions that satisfy the needs of our gravity optimization problem. We apply a diffusion factor (using the \textit{lift} operation) between consecutive gravity variables in our pose graph to account for possible odometry drift.}
    \label{fig:s2}
    \vskip -0.2in
\end{figure}

The standard gravitational acceleration on Earth ($9.80665\text{ m}/\text{s}^2$) is accurate to high precision across Earth's surface ($\pm 0.3\%$), and local values are known where higher precision is required. Therefore, estimation can benefit from restricting gravity to the unit sphere $S^2$, a smooth manifold, scaled by this known constant. While there is more than one approach to optimizing on a smooth manifold, GTSAM takes advantage the Lie group structure where it exists \cite{kaess2011isam2}. The use of Lie groups is partially motivated by a need to represent the (Gaussian) covariance of optimized variables, and in this case they can readily be expressed in the tangent space of the manifold, where values are mapped to the manifold via the Lie group's exponential map. While commonly used state spaces in robotics such as $SE(3)$ for 3D pose are Lie groups, $S^2$ is not; however, we can still construct functions on $S^2$ that satisfy the needs of our optimization problem (Fig.~\ref{fig:s2}).

First, we need a logarithmic and exponential map to and from the tangent space at any point on the manifold, since the optimizer will manipulate the variable and its covariance in the tangent space. (\ref{eq:exp}) and (\ref{eq:log}) are presented by \cite{dubbelman2011intrinsic} as the exponential and logarithmic maps at the origin.
\begin{align}
    \text{Exp}_{\id}(\tanEl) &= \cos(\|\tanEl\|)\id + \sin(\|\tanEl\|)\tanB_\id\frac{\tanEl}{\|\tanEl\|}, \quad &&\tanEl \in \mathbb{R}^2 \label{eq:exp} \\
    \text{Log}_{\id}(\manEl) &= \arccos(\id \cdot \manEl)\frac{\tanB_\id\top\manEl}{\|\tanB_\id^\top\manEl\|} , \quad &&\manEl \in S^2 \label{eq:log}
\end{align}
Here, one must select an identity element or origin $\id \in S^2 \subset \mathbb{R}^3$, as well as an orthonormal basis for its tangent space $\tanB_{\id} \in \mathbb{R}^{3 \times 2}$; a natural choice is $\id = \mathbf{e}_3$, $\tanB_{\id} = \left[\mathbf{e}_1\;\mathbf{e}_2\right]$. Additionally, $\text{Exp}_{\id}(0) = \id$ and $\text{Log}_{\id}(\id) = 0$ are defined to maintain continuity. Note that $\text{Log}_{\id}(-\id)$ is undefined. To obtain the appropriate maps about any point (except $-\id$), a rotation (\ref{eq:r}) is defined for each point $\manEl \in S^2$, $\manEl \neq -\id$ \cite{dubbelman2011intrinsic}.
\begin{align}
    \R(\manEl) = \I + \left[{\id \times \manEl}\right]_\times + \frac{1}{1+({\id \cdot \manEl})}\left[{\id \times \manEl}\right]_\times^2 \label{eq:r}
\end{align}
The angle of rotation in (\ref{eq:r}) is $\theta = \arccos{(\id \cdot \manEl)}$ and the axis is $\hat{r} = \frac{1}{\sin{\theta}}(\id \times \manEl)$. This is the shortest rotation that takes the vector $\id$ to $\manEl$, and it is well known that this rotation is not unique when $\manEl = -\id$. Luckily, we can ensure that we never evaluate $\text{Log}_{\id}(\cdot)$ or $\R(\cdot)$ near $-\id$ in our optimization problem. Finally, the mappings (\ref{eq:retract}) and (\ref{eq:local}) are defined \cite{dubbelman2011intrinsic}, which we set as the \textit{retract} and \textit{local} (a.k.a. lift) operations for $S^2$ in GTSAM, respectively \cite{sola2018lie}.
\begin{align}
    \text{Exp}_{\manEl_1}(\tanEl_2) &= \R(\manEl_1)\text{Exp}_{\id}(\tanEl_2) \label{eq:retract} \\
    \text{Log}_{\manEl_1}(\manEl_2) &= \text{Log}_{\id}\left(\R(\manEl_1)^{-1}\manEl_2\right) \label{eq:local}
\end{align}
An insight here is that defining $\manEl_1 \circ \manEl_2 \triangleq 
\R(\manEl_1)\manEl_2$ yields an operation analogous to a group operation, although we cannot call it as such because it is not associative.
In the factor graph, each factor is characterized by an error function which must have an associated Jacobian to inform the optimizer. For the factors involving gravity, we use the definition of the right Jacobian for Lie groups as described in \cite{sola2018lie}, as it represents perturbations in the local tangent space.

\subsubsection{Diffusion Factor}
We apply a diffusion factor between consecutive gravity variables to account for possible odometry drift. The error (\ref{eq:e_diff}) uses the logarithmic map so that its magnitude is equal to the length of the geodesic between the two variables.
\begin{align}
    e_\text{d}(\manEl_1, \manEl_2) = \text{Log}_{\manEl_1}(\manEl_2) = \text{Log}_{\id}\left(\R(\manEl_1)^{-1}\manEl_2\right) \label{eq:e_diff}
\end{align}
By assuming that the estimates $\manEl_1$ and $\manEl_2$ are always close to each other ($\leq 5^\circ$), the Jacobians can be approximated quite well by (\ref{eq:J_diff}), which is sufficient to make optimization work.
\begin{align} 
    \frac{De_\text{d}(\manEl_1, \manEl_2)}{D\manEl_1} \approx -\I_{2 \times 2}, \quad \frac{De_\text{d}(\manEl_1, \manEl_2)}{D\manEl_2} &\approx \I_{2 \times 2} \label{eq:J_diff}
\end{align}
One may improve the Jacobian approximation by taking into account the Jacobian of $\text{Log}_{\id}(\manEl)$ and using the chain rule. This is presented in (\ref{eq:J_Log}) where $v \defeq \tanB_\id^\top\manEl$.
\begin{multline}
    \frac{D\text{Log}_{\id}(\manEl)}{D\manEl} = \\
    \frac{1}{\|v\|^2}
    \begin{bmatrix}v_1 & -v_2 \\ v_2 & v_1\end{bmatrix}
    \begin{bmatrix}1 & 0 \\ 0 & \frac{\arccos{(\id \cdot \manEl)}}{\|v\|}\end{bmatrix}
    \begin{bmatrix}v_1 & -v_2 \\ v_2 & v_1\end{bmatrix}^\top
  \label{eq:J_Log}
\end{multline}

\subsubsection{Prior Factor}
The prior factor is used to constrain the initial gravity estimate given an expected value $\bar{\manEl}$ and covariance. The error function $e_\text{p}(\manEl) = -\text{Log}_{\manEl}\left(\bar{\manEl}\right)$ is similar to that of the diffusion factor in (\ref{eq:e_diff}), and thus the Jacobian can be approximated in the same way.

\subsubsection{Odometry Factor}
Section \ref{sec:o_factor} discusses the odometry factor, whose error function is the left-hand side of (\ref{eq:o_factor}). To find the Jacobian with respect to the gravity estimate $\manEl$, we only need to consider the term $\mathrm{M}_{\mathrm{g}}\mathbf{g}_\mathcal{I} = \|\mathbf{g}_\mathcal{I}\|\mathrm{M}_{\mathrm{g}}\manEl$. Noting that the standard definition of the Jacobian would yield $\mathbf{J} = \|\mathbf{g}_\mathcal{I}\|\mathrm{M}_{\mathrm{g}}$, the on-manifold Jacobian (with respect to perturbations in the tangent space) is then $\|\mathbf{g}_\mathcal{I}\|\mathrm{M}_{\mathrm{g}}\tanB_\manEl$, where $\tanB_\manEl = \R(\manEl)\tanB_\id$.

\subsection{Factor Graph for Intrinsics \& Gravity}

The composition of the factor graph straightforward, and is seen in Fig. \ref{fig:factor_graph}. We choose a duration $T_m$ during which the accelerometer instrinsics and gravity are assumed to remain constant. We then create variables $\mathbf{S}_0$, $\mathbf{b}_0$, $\mathbf{g}_0$ which correspond to the interval $\left[0, T_m\right)$, then $\mathbf{S}_1$, $\mathbf{b}_1$, $\mathbf{g}_1$ for the interval $\left[T_m, 2T_m\right)$, and so on. All factors are given as Gaussian distributions, including the odometry factors whose error functions are given by the left-hand side of (\ref{eq:o_factor}) and covariances given by (\ref{eq:o_factor_cov}). We also start with a prior factor on $\mathbf{S}_0$, $\mathbf{b}_0$, $\mathbf{g}_0$ representing \textit{a priori} knowledge. This knowledge includes the fact that $\textbf{S} \approx I$ and $\textbf{b} \approx 0$, roughly speaking, so reasonable standard deviations are chosen; $\textbf{g}_0$ is loosely constrained, as it can be approximated by the first few IMU measurements (if the robot is static on startup).

Next, a ``diffusion" factor is added between consecutive variables to describe the random walk for each quantity. To roughly model random fluctuations and the effects of temperature change, we assume that each component of $\mathbf{S}$ and $\mathbf{b}$ changes over time according to a Wiener process. For bias, the error functions are of the form $\mathbf{b}_{i+1} - \mathbf{b}_{i}$ and each factor has covariance $\sigma^2T_m\I$, and the same is done for sensitivity represented as a vector. The $\sigma$ for sensitivity and bias are chosen to approximate the expected behavior of the accelerometer. The random walk for gravity is modeled differently: it is expected to drift slightly with each keyframe placement, so we similarly model it as a random walk but with variance of $\mathbf{g}_{i+1} - \mathbf{g}_{i}$ scaled by the number of keyframes placed during the time interval associated with $\mathbf{g}_{i}$.

Subsequent variables for the accelerometer intrinsics get an additional prior-like factor in order to prevent the estimates from deviating unreasonably from $\mathbf{S} \approx \I$ and $\mathbf{b} \approx 0$. This accounts for the fact that a random walk does not perfectly model the evolution of the intrinsics, since they appear to be bounded. The variance for these factors was tuned by experiment using synthetically-generated datasets with known intrinsics. Estimation and marginalization are executed by iSAM2 \cite{kaess2011isam2}, with relinearization performed at every step. Variables are marginalized out when they are older than the lag time $T_l \gg T_m$. While a new factor is created after every pose measurement, we choose to recompute estimates less frequently to save CPU time.

\begin{figure}
    \centering
    \vspace{2mm}
    \includegraphics[width=0.9\columnwidth]{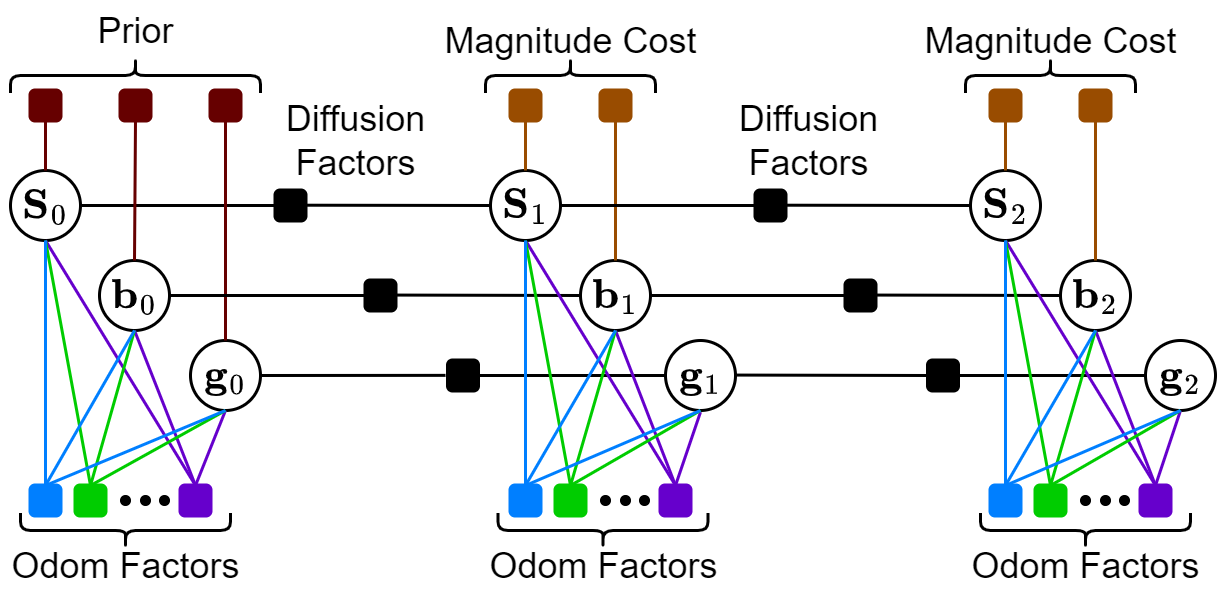}
    \vspace{-2mm}
    \caption{Factor graph structure used to estimate gravity and accelerometer intrinsics. Prior factors constrain the initial estimates to reasonable values. Odometry factors are added after each new pose measurement. A separate set of variables is added for every time interval of length $T_m$. Diffusion factors describe the random walk of each variable. Magnitude cost factors prevent the sensitivity and bias estimates from becoming unreasonable. }
    \label{fig:factor_graph}
    \vskip -0.2in
\end{figure}

\subsection{Gravity Factors for Mapping}\label{sec:grav_factor} 
Another type of factor constrains the direction of gravity in the body frame for each keyframe.
When a new keyframe is placed, the current estimate $\manEl_\mathcal{O}$ and covariance $\Sigma_\mathcal{O}$ for gravity are captured, where $\manEl_\mathcal{O}$ denotes the \textit{direction} of gravity as a unit vector in the inertial frame as seen by the \textit{odometry} module. To take the gravity estimate (mean) and covariance from the inertial frame to the body frame, we use
\begin{align}
    \bar{\manEl}_\mathcal{B} &= \R_{\mathcal{O}}^{-1}\bar{\manEl}_\mathcal{O} \,, \\
    \Sigma_\mathcal{B} &= \left(\tanB_{\bar{\manEl}_\mathcal{B}}\top \R_{\mathcal{O}}^{-1} \tanB_{\bar{\manEl}_\mathcal{O}}\right) \Sigma_\mathcal{O} \left(\tanB_{\bar{\manEl}_\mathcal{B}}\top \R_{\mathcal{O}}^{-1} \tanB_{\bar{\manEl}_\mathcal{O}}\right)^\top .
\end{align}
Unlike in the odometry module where the inertial frame is expected to drift, we enforce a global direction of gravity $\manEl_\mathcal{M} = \mathbf{e}_3$ in the map frame $\mathcal{M}$. Given a robot orientation in the map frame $\R_\mathcal{M}$, the direction of gravity in the body frame is then $\R_\mathcal{M}^{-1}\manEl_\mathcal{M}$. For each keyframe pose in the pose graph with orientation $\R_\mathcal{M}$, the error of the new factor is $e_\text{g}(\R_\mathcal{M}) = \text{Log}_{\bar{\manEl}_\mathcal{B}}\left(\R_\mathcal{M}^{-1}\manEl_\mathcal{M}\right)$, with covariance $\Sigma_\mathcal{B}$. The on-manifold Jacobian of the error can be derived similarly to those in Section \ref{sec:s2}, but now the input space is $SO(3)$ and the Jacobian of the inverse rotation is used.


\section{Results}
We demonstrate the effectiveness of our approach via a map optimization module (integrated into DLIO \cite{dlio}) that constrains relative poses between keyframes and the alignment to gravity as outlined in Section \ref{sec:grav_factor}. We compare DLIO with the new gravity and intrinsics estimator (w/ GE) against DLIO as presented in \cite{dlio} (w/o GE), as well as two state-of-the-art LIO methods, FAST-LIO2 \cite{xu2022fast} and D-LIOM \cite{wang2022dliom}. DLIO on its own does not estimate gravity, but estimates the state along with gyroscope and accelerometer biases using a geometric observer \cite{dlio}; the proposed estimator replaces only the accelerometer bias estimation in DLIO, keeping all else the same. FAST-LIO2 estimates gravity via a probabilistic filter which jointly optimizes all state variables \cite{xu2022fast}, while D-LIOM estimates gravity as a self-contained step in its local trajectory optimization \cite{wang2022dliom}. Note also that FAST-LIO2 does not perform global map optimization, while D-LIOM does. However, D-LIOM only incorporates gravity constraints into its local optimization. Additionally, all these algorithms include a static gravity calibration upon initialization which assumes zero accelerometer bias; for DLIO, this calibration has a duration of 1s. In these experiments, we used both public benchmark \cite{ramezani2020newer, zhang2021multicamera} and self-collected datasets, including: (A) Slow Yaw; (B) Lab-E4-Boelter; (C) Dynamic Spinning; (D) Maths{-}Hard; (E) Stairs; (F) Quad with Dynamics; (G) Parkland; and (H) Cloister. The self-collected dataset (A) is discussed in Section \ref{sec:imu_prop} and used an Ouster OS0 IMU, while (B) is a loop around campus buildings including hallways and transitions between indoors and outdoors and used an MPU-6050 IMU.

\subsection{Computation Time}
 All experiments were conducted on a 16-core Intel i7-11800H CPU. In the implementation, we choose to perform any pending marginalization and optimization of the factor graph once every 5 LiDAR scans (i.e., every 0.5s for 10Hz and every 0.25s for 20Hz LiDAR), and this runs in its own thread. We report an average computation time of 3-4ms per factor graph update for 10Hz LiDAR, and 4-5ms for 20Hz LiDAR. This indicates extremely little added CPU usage.


\subsection{IMU Intrinsics and Propagation Accuracy}\label{sec:imu_prop}

\begin{table}[!t]
\centering
\vspace{2mm}
\setlength{\tabcolsep}{8 pt}
\caption{IMU Propagation Accuracy}
\vspace{-2mm}
\begin{tabular}{ |l||c|c|c|c|c| }
 \hline
 \multirow{2}{*}{Method} & \multicolumn{5}{c|}{RMSE IMU-LiDAR Deviation [cm]} \\
 \cline{2-6}
 & A & B & C & D & E \\
 \hline
 DLIO (w/o GE) \cite{dlio} & 1.9 & 8.7 & 4.6 & 10.3 & 4.6 \\
 \hline
 DLIO (w/ GE) & \textbf{1.5} & \textbf{7.6} & \textbf{3.5} & \textbf{5.1} & \textbf{2.6} \\
 \hhline{|=||=|=|=|=|=|}
 \% improvement & 21\% & 13\% & 24\% & 50\% & 43\%\\
 \hline
\end{tabular}
\label{tab:dp}
\end{table}

\begin{figure}
    \centering
    \vspace{-4mm}
    \includegraphics[width=0.95\columnwidth]{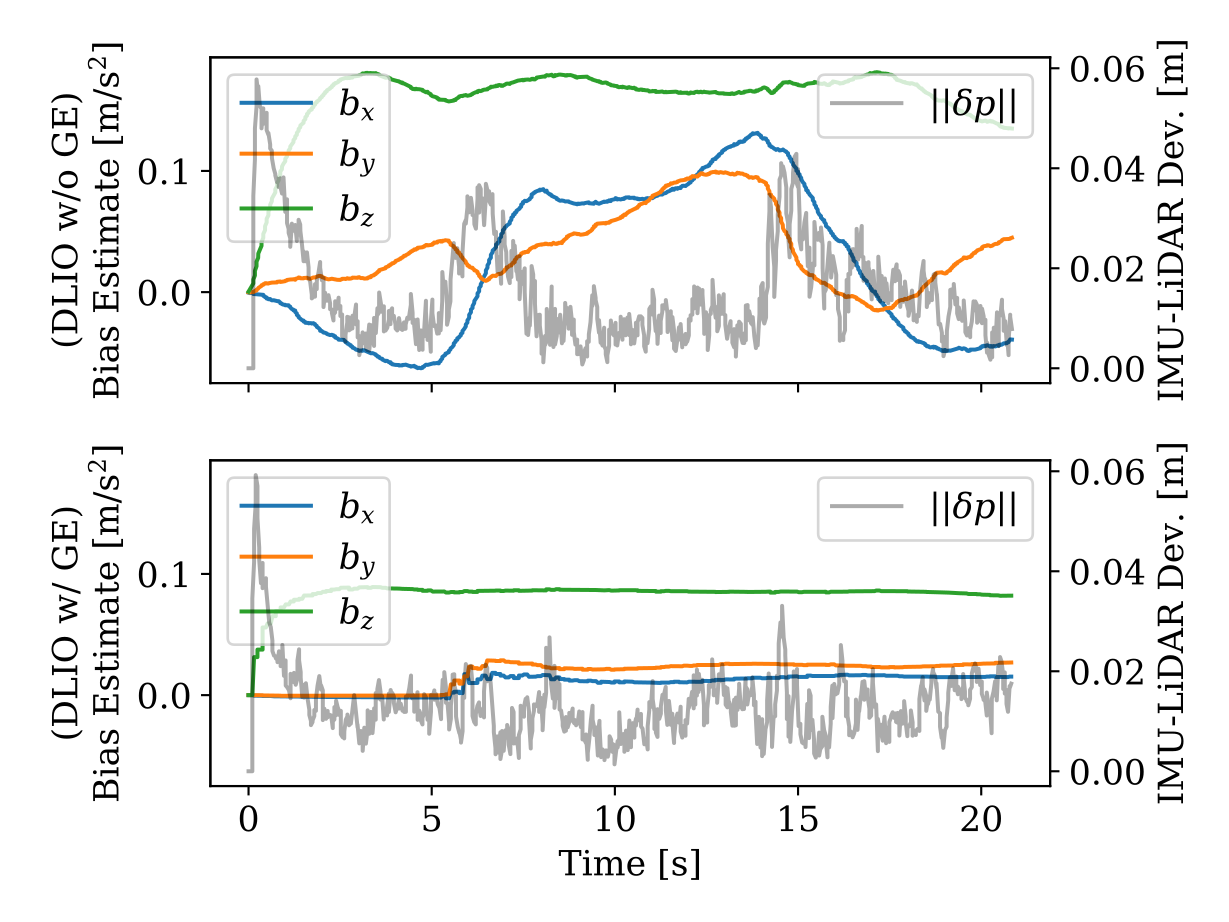}
    \vspace{-4mm}
    \caption{Bias estimates and IMU-LiDAR deviation for DLIO w/o GE \cite{dlio} (top) and DLIO w/ GE, i.e. with the proposed estimator (bottom), on the Slow Yaw dataset. IMU-LiDAR deviation is the distance between the current LiDAR pose and the IMU prediction starting from 0.5s in the past. This demonstrates the shortcomings of assuming constant gravity, as bias estimates fluctuate and IMU error spikes (not present in DLIO w/ GE).}
    \label{fig:bias}
    \vskip -0.2in
\end{figure}

We first tested our estimator on a custom synthetic dataset (with ground truth intrinsics) where the sensor platform moves inside a box with piecewise-constant linear jerk and angular acceleration trajectories. Point cloud and IMU measurements are perturbed with Gaussian noise, constant IMU biases, and a constant accelerometer sensitivity matrix. Using simulated trajectories with good excitation, the average error for the accelerometer instrinsics at the end of trajectories is 0.09 m/s$^2$ for biases and 0.008 (0.8\%) for components of the sensitivity matrix. When biases are especially detrimental to the initial gravity calibration, the reduction in bias error compared to DLIO w/o GE is up to 36\%.

We also tested the approach on a self-collected Slow Yaw dataset with an Ouster OS0 LiDAR, consisting of two $180^\circ$ rotations with stopping in between. 
Fig.~\ref{fig:bias} shows that the bias estimates from DLIO's geometric observer~\cite{dlio} highly depend on the yaw angle due to the error in the initial gravity calibration, while the new method updates the estimates to be consistent with the entire history of measurements, and thus they remain stable after the first rotation. This also yields a more stable IMU-LiDAR deviation.
In Table \ref{tab:dp}, we show the RMSE IMU-LiDAR deviation for several datasets with varying IMU sensors. The IMU-LiDAR deviation is defined as the distance between a LiDAR position measurement and the IMU prediction propagated from some time in the past, in this case 0.5s. This is a good proxy for the true IMU prediction error since the LiDAR estimates are generally much more accurate. This metric is important because IMU propagation may be relied upon more heavily when observability from other sensors is low, e.g. due to poorly constrained point cloud registration.
Consistent improvement is seen using the new estimator compared to DLIO w/o GE.

\subsection{Gravity Estimation Accuracy}





\begin{table}[!t]
\centering
\vspace{2mm}
\setlength{\tabcolsep}{8 pt}
\caption{Gravity Vector Accuracy}
\vspace{-2mm}
\begin{tabular}{ |l||c|c|c|c|c| }
 \hline
 \multirow{2}{*}{Method} & \multicolumn{5}{c|}{Final Gravity Direction Error [degrees]} \\
 \cline{2-6}
 & D & E & F & G & H \\
 \hline
 DLIO (w/o GE) \cite{dlio}    & 1.31 & 1.08 & 1.93 & 2.99 & 1.31 \\
 \hline
 D-LIOM \cite{wang2022dliom}  & 3.81 & Failed & 1.62 & 2.10 & 13.37 \\
 \hline
 FAST-LIO2 \cite{xu2022fast}  & \textbf{0.14} & \textbf{0.29} & 0.39 & \textbf{0.14} & \textbf{0.06} \\
 \hline
  DLIO (w/ GE\raisebox{0.15ex}{{$\ominus$}}S)              & 0.39 & 0.49 & 0.04 & 0.35 & 0.12 \\
 \hline
 DLIO (w/ GE)                 & 0.37 & 0.31 & \textbf{0.02} & 0.32 & 0.07 \\
 \hline


 %
\end{tabular}
\label{tab:grav}
\end{table}

\begin{figure}[t!]
    \centering
    \vspace{-2mm}
    \includegraphics[width=0.99\columnwidth]{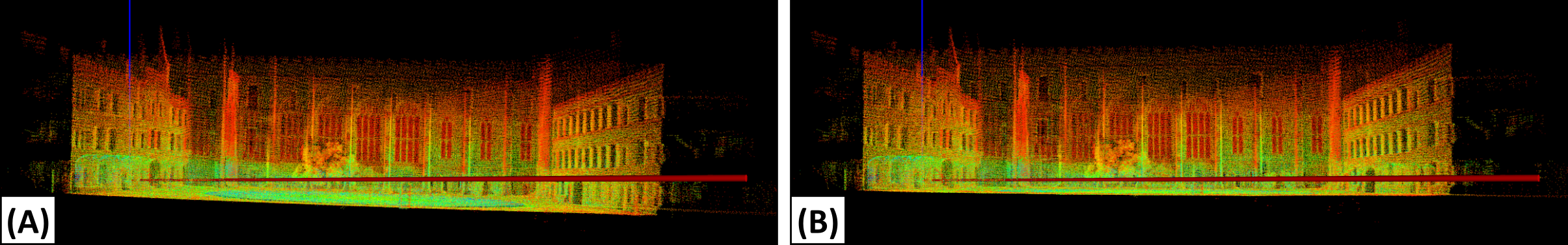}
    \vspace{-6mm}
    \caption{The advantage of estimating gravity can be visualized in the above figure using the Newer College - Quad with Dynamics dataset. The error of initial gravity calibration (DLIO w/o GE) is clearly visible (A) while the gravity-corrected map (DLIO w/ GE) is far better aligned (B).}
    \vskip -0.2in
    \label{fig:map_tilt}
\end{figure}


Next, we compared the final gravity estimate of several trajectories to the vertical unit vector $\mathbf{e}_3 = [0,\,0,\,1]^\top$ after alignment with evo \cite{grupp2017evo}. More precisely, we generated the $\text{SE}(3)$ alignment that best aligned each trajectory with the ground truth, then applied the resulting rotation to the gravity estimate (produced in the odometry frame) and measured the angle to $\mathbf{e}_3$. We can reasonably assume that the Newer College ground truth trajectories are aligned with gravity due to the use of the high-grade Leica BLK360 laser scanner to produce the ground truth maps, which has an IMU and an auto-tilt feature (limited details available) \cite{ramezani2020newer}.

These results are presented in Table \ref{tab:grav} (also see Fig.~\ref{fig:map_tilt}), where the proposed method (DLIO w/ GE) achieves an accuracy comparable with FAST-LIO2 \cite{xu2022fast}.
Although FAST-LIO2 with its full-state IEKF \cite{xu2022fast} achieves the best gravity vector estimation accuracy for the tested datasets, the integration of the proposed approach with DLIO yields more accurate maps, highlighting the benefits of a decoupled approach.
We additionally compared the accuracy of the 60-second optimization window used for our method in Table \ref{tab:grav} to a 10-second optimization window and found that the longer window yielded a 21\% improvement on average, lending credence to the argument for a longer optimization window. 
Furthermore, our decision to estimate accelerometer sensitivity is supported by the improvement (28\% on average) of the proposed (DLIO w/ GE) over the proposed without sensitivity estimation (DLIO w/ GE\raisebox{0.15ex}{{$\ominus$}}S).
Finally, note that DLIO w/o GE represents the error when assuming a constant gravity vector after initial calibration, so in theory any algorithm which estimates gravity online should perform better. Interestingly, this is not always the case for D-LIOM, which may be partly due to the lack of \textit{joint} estimation of biases and gravity.

\subsection{Mapping Accuracy}


Finally, we compared the mapping accuracy between each method by computing a cloud-to-cloud error between each algorithm's output map and the ground truth. Specifically, we used the Stairs, Cloister, and Short Experiment datasets from Newer College \cite{ramezani2020newer, zhang2021multicamera} to comprehensively evaluate accuracy in small, medium, and large environments, respectively. Cloud-to-cloud error was computed using CloudCompare~\cite{cloudcompare} after manual alignment with varying maximum distance thresholds (i.e., $0.1, 0.5, 1.0$) depending on the size of the environment; this was to ensure a fair comparison between all algorithms by evaluating only points within the ground truth map. Ground truth was provided by \cite{ramezani2020newer, zhang2021multicamera}.

As seen in Table~\ref{tab:map}, our optimization-based method provided the lowest cloud-to-cloud error as compared to all other methods. Additionally, we computed an end-to-end drift using the (B) dataset and observed the lowest error from our method (0.22m) as compared to D-LIOM (8.59m), FAST-LIO2 (2.14m), and DLIO w/o GE (1.07m). 
This can additionally be seen in Fig.~\ref{fig:map} using the Hilti-Oxford Cupola \cite{zhang2023hiltioxford} dataset, in which extreme drift occurred due to the challenging localization within the dataset's many staircases, but our method was able to compensate and correct for this.





\begin{table}[!t]
\centering
\vspace{2mm}
\setlength{\tabcolsep}{12 pt}
\caption{Mapping Accuracy}
\vspace{-2mm}
\begin{tabular}{ |l||c|c|c|c| }
 \hline
 \multirow{2}{*}{Method} & \multicolumn{3}{c|}{Cloud-to-Cloud Error [m]} \\
 \cline{2-4}
 & Stairs & Cloister & Short  \\
 \hline
 DLIO (w/o GE) \cite{dlio} & 0.0535 & 0.2608 & 0.1372 \\
 \hline
 D-LIOM \cite{wang2022dliom} & Failed & 0.3572 & 0.4312 \\
 \hline
 FAST-LIO2 \cite{xu2022fast} & 0.0667 & 0.3556 & 0.4135 \\
 \hline
 DLIO (w/ GE) & \textbf{0.0488} & \textbf{0.1791} & \textbf{0.1334}\\
 \hline
\end{tabular}
\vskip -0.2in
\label{tab:map}
\end{table}


\section{Discussion}
This work presented a fixed-lag optimizer for estimating gravity and accelerometer intrinsics via a factor graph, which can be integrated into any odometry or SLAM system.
This method is unique in jointly optimizing intrinsics and gravity over a large time window, and is designed to maximize accuracy when excitation and observability are limited.
Accelerometer scale factor is also estimated. 
The estimator enables better IMU prediction compared to a system that assumes a constant gravity vector, and more accurate maps compared to state-of-the-art SLAM algorithms which do not employ gravity-based global map correction. 
Gravity estimates are competitive with the state-of-the-art, with the added benefit of lower computational burden and greater freedom to integrate with any existing odometry or SLAM algorithm.
Future work includes accounting for uncertainty in orientation estimates within the estimator, or perhaps taking into account other measurement uncertainties such as gyroscope scale factor. 
Also, additional analysis should be done in regards to the measurement model for intrinsics and gravity, such as one that leverages velocity information.


\vspace{2mm}
\noindent \small \textbf{Acknowledgements:} The authors would like to thank Aaron John Sabu for their help with figure generation.

\bibliographystyle{IEEEtran}
\bibliography{IEEEbib}

\end{document}